\documentclass[submission,copyright,creativecommons]{eptcs}
\providecommand{\event}{FROM 2026} 
\usepackage{iftex}

\ifpdf
  \usepackage{underscore}         
  \usepackage[T1]{fontenc}        
\else
  \usepackage{breakurl}           
\fi

\usepackage{graphicx}%
\usepackage{multirow}%
\usepackage{amsmath,amssymb,amsfonts}%
\usepackage{amsthm}%
\usepackage{mathrsfs}%
\usepackage[title]{appendix}%
\usepackage[table]{xcolor}%
\usepackage{textcomp}%
\usepackage{manyfoot}%
\usepackage{booktabs}%
\usepackage{algorithm}%
\usepackage{algorithmicx}%
\usepackage{algpseudocode}%
\usepackage{listings}%
\usepackage{caption}
\usepackage{xspace}
\usepackage{cleveref}
\usepackage[normalem]{ulem}
\usepackage[table]{xcolor}
\usepackage{array}
\usepackage{graphicx}
\usepackage{cleveref}
\usepackage{wrapfig}
\usepackage{xspace}

\usepackage{booktabs}
\usepackage{multirow}
\usepackage{siunitx}

\usepackage{tabularx}
\usepackage{array}

\definecolor{keywordcolor}{RGB}{0,0,180}
\definecolor{commentcolor}{RGB}{0,120,0}
\definecolor{stringcolor}{RGB}{160,0,0}
\definecolor{ghostcolor}{RGB}{140,80,180}

\lstdefinelanguage{Verus}{
  language     = {[Sharp]C},
  morekeywords = {fn,pub,struct,impl,let,mut,if,else,for,while,return,
                  true,false,Some,None,Option,u32,usize,bool,nat,int,
                  requires,ensures,invariant,decreases,proof,spec,exec,
                  assume,assert,by,open,closed,Ghost,Set,Seq,Map,
                  verus,forall,exists,old,ghost,trigger,recommends},
  keywordstyle = \color{keywordcolor}\bfseries,
  comment      = [l]{//},
  commentstyle = \color{commentcolor}\itshape,
  stringstyle  = \color{stringcolor},
  basicstyle   = \ttfamily\scriptsize,
  breaklines   = true,
  keepspaces   = true,
  columns      = flexible,
  showstringspaces = false,
  literate     = {&&\&}{{\&\&\&}}3 {==>}{{==>}}3 {<==>}{{<==>}}4,
}

\definecolor{typecolor}{HTML}{006699}		
\definecolor{keywordspeccolor}{HTML}{d94a7a}		
\colorlet{membercolor}{blue}
\colorlet{keywordcolor}{blue}

\lstdefinelanguage{trait}{
  morekeywords = [1]{Input, Output, Return},
  morekeywords = [2]{push_back, push_front, self, first, last, next, prec, value, is_empty, self', first', last', next', prec', value', is_empty', insert_before, insert_after},
  morekeywords = [3]{Value, T, NodeId, usize, u32, NonNull, W, R, TheAlloc, A},
  morekeywords = [4]{if, then, else},
  morekeywords = [5]{Option, Slab, pub, struct, type, PhantomData, Vec, Arena, mut, Rc, Weak, RefCell, Box, Allocator, Global},
  keywordstyle = [1]\footnotesize\sffamily\color{keywordspeccolor},
  keywordstyle = [2]\ttfamily\color{membercolor},
  keywordstyle = [3]\ttfamily\color{typecolor},
  keywordstyle = [4]\bfseries,
  keywordstyle = [5]\ttfamily\color{blue},
  sensitive = true,
  morecomment = [l]{//},
  morecomment = [s]{/*}{*/},
  morecomment = [s]{/**}{*/},
  commentstyle = \color{gray},
  morestring = [b]",
  stringstyle = \color{purple}
}
\Crefname{listing}{Specification}{Specifications}

\DeclareCaptionFormat{listing}{\par\vskip1pt#1#2#3}
\newcommand{\None}{\ensuremath{\mathit{None}}\xspace}

\newcommand{\pred}{{\tt\textcolor{membercolor}{prec}}}
\newcommand{\suc}{{\tt\textcolor{membercolor}{next}}}

\newcommand{\id}{\ensuremath{\mathit{id}}\xspace}

\newcommand{\wf}{\texttt{wf()}}
\newcommand{\fullwf}{\texttt{full\_wf()}}
\newcommand{\chainwf}{\texttt{chain\_wf()}}
\newcommand{\NodeId}{\texttt{NodeId}}

\newcommand{\code}[1]{\texttt{#1}}

\title{Fewer Assumptions by Design: A Reusable Skill for LLM-Assisted Verus Verification}
\author{Andrada-Livia Antoneac$^{1,2}$
\email{aantoneac@bitdefender.com}
\and
Dorel Lucanu$^{2}$
\email{dorel.lucanu@info.uaic.ro}
\and 
Dragoș Teodor Gavriluț$^{1,2}$
\email{dgavrilut@bitdefender.com}
\institute{${}^{1}$Bitdefender, \quad {}$^{2}$Alexandru Ioan Cuza University of Iași}
}
\def\titlerunning{Fewer Assumptions by Design: A Reusable Skill for LLM-Assisted Verus Verification}
\def\authorrunning{A.L. Antoneac, D. Lucanu \& D.T. Gavriluț}
\begin{document}
\maketitle

\begin{abstract}
LLM-assisted Verus verification is a less tedious method to verify Rust implementations, but paired with self-referential structures, e.g., Doubly Linked Lists (DLLs)---notoriously difficult to formalise for verification---it becomes a substantially more demanding verification task.   
Moreover, a specification weakness can arise when verification relies on unproven or invalidated assumptions, such as axiomatic lemmas and assume statements. We investigate whether LLM agents can synthesize strong DLL specifications while minimizing these trusted base. The analysis follows three different approaches: manual verification, property-specific verification, and a defined skill for the specific case of DLLs and certain properties of this type of data structure. The skill encodes domain knowledge and a task-decomposition strategy. We show that an LLM agent equipped with a carefully designed verification skill can generate strong, low-trust specifications for DLLs in Verus.
\end{abstract}

\section{Introduction}
\label{sec:intro}
Verifying multiple implementations against a shared abstract specification is a good stress test for LLM-assisted formal verification. The proof obligations share a common logical skeleton, but each has unique requirements that must be handled concretely.

The paper addresses the question of \emph{whether an AI model can be used to efficiently verify a family of structurally related Rust implementations against a shared formal specification, and whether the process can be codified into a reusable methodology}.

DLLs are a canonical data structure, conceptually simple, but notoriously difficult to verify formally because their correctness depends on mutual consistency constraints between forward and
backward reference chains. In~\cite{dll-paper}, eleven different Rust implementations of a DLL are compared in terms of execution time, memory footprint, and memory safety. All implementations have the same abstract trait-based description, comprising eight operations (\code{next}, \code{prec}, \code{push\_back}, \code{push\_front},
\code{insert\_before}, \code{insert\_after}, \code{delete}, \code{search}), and differ in memory mode (heap-based, resizable-array-based, and map-based) and in the node identifier type. That paper evaluates the implementations
empirically for time and memory performance, and tests them for three
safety vulnerabilities.  What it does not do is formally verify them.
What we aim to add here is a way to formally verify these implementations, as their functional correctness has not been addressed there.

Verifying all eleven implementations against the shared abstract specification
would be an ideal case study for answering the main research question and the work presents the first steps we take to achieve this goal. 

The main contributions include a systematization of the challenges raised by such a verification, a set of principles for designing a verification-oriented skill for AI agents based on these challenges, and the implementation of a skill based on these principles.

The article is structured as follows: \Cref{sec:preliminariesAndRelatedWork} presents the use-case and the similar approaches to AI-assisted verification, \Cref{sec:verificationCaseStudy} is a progressive, three-stage case study of formal verification of DLL and the resulting challenges, \Cref{sec:skill} presents the reusable verification generation skill we created for this context, found at \url{https://github.com/andradaAntoneac/generate-verus-specification}, \hfill \break along with key findings from its results, summarized in the concluding \Cref{sec:conclusions}.

\section{Background}
\label{sec:preliminariesAndRelatedWork}
\subsection{Formal Verification in Rust}
The most popular tools for formal verification in Rust can be grouped by the methodology they follow, showcasing the differences between them. \emph{Model checkers} explore program execution paths rather than construct a mathematical proof, performing an exhaustive search to check all possible states a program can reach and verify whether a property holds for those states \cite{jhala2009software}; they provide high automation but limited capabilities, as exemplified by \emph{Kani} \cite{vanhattum2022verifying}, a tool built over CBMC \cite{kroening2014cbmc}, a bit-precise bounded model checker designed for C programs and ideal for catching memory-safety violations such as use-after-free and out-of-bounds accesses in unsafe Rust. \emph{Deductive verifiers}, in contrast, explore all possible execution paths and all possible inputs to construct a mathematical proof, which makes them unbounded and able to provide full correctness proofs, usually relying on an SMT prover; representatives include \emph{Prusti} \cite{astrauskas2022prusti}, built over the Viper framework (Verification Infrastructure for Permission-Based Reasoning) yet still in prototype form, and \emph{Verus} \cite{verus}, which uses Z3 directly, offers a specification language very similar to Rust, and is designed specifically for low-level systems code. Finally, \emph{translation-based} approaches translate Rust code into another formal representation that can be verified independently, usually a backend like Why3, Coq, Lean, or F*; these include \emph{Creusot} \cite{denis2022creusot}, which translates Rust into WhyML for verification via Why3, \emph{Aeneas} \cite{ho2022aeneas}, which translates safe Rust into a pure functional model for proof assistants (Coq, Lean, F*), and \emph{RustBelt} \cite{jung2017rustbelt}, which provides the theoretical justification for why safe abstractions over unsafe code are sound but is not suitable for proofs over specific cases such as the one presented in this article.

\subsection{AI-Assisted Verification}
Using large language models (LLMs) as assistants for formal verification is a rapidly evolving
research direction.  Most existing work automates the proof-engineering pipeline end-to-end,
treating the LLM as a component in a larger system.

\paragraph{Automated pipeline approaches for Verus.}
\emph{AutoVerus}~\cite{autoverus} is an early automated system from Microsoft Research that
orchestrates a network of LLM agents mimicking the three human proof construction phases:
preliminary proof generation, refinement guided by generic tips, and debugging guided by
verification errors.  The same group later introduced a \emph{self-evolution} strategy in which
GPT-4o is used to synthesise proof annotations for over one thousand Rust functions, and the
resulting data is used to fine-tune a smaller open-source model that iteratively improves
itself~\cite{autoverus}.
\emph{VeriStruct}~\cite{veristruct}, accepted at TACAS~2026, extends AI-assisted verification
from single functions to full data-structure modules in Verus.  Its planner module
orchestrates the systematic generation of view functions, type invariants, specifications, and
proof code; a subsequent repair stage corrects annotation errors automatically.
Applied to eleven Rust data-structure modules, VeriStruct succeeds on ten,
verifying 128 of 129 functions~(99.2\%).
\emph{VeruSAGE}~\cite{verusage} introduced a system-verification benchmark of 849 proof
tasks extracted from eight open-source Verus-verified Rust projects, and studied how different
LLMs (Claude Sonnet~4, Sonnet~4.5, GPT-5, and o4-mini) perform across specialised agents
handling logic, arithmetic, and proof context.  The best LLM–agent combination solves over
80\% of the benchmark tasks.

\paragraph{Automated pipeline approaches for Dafny.}
Parallel work on the Dafny verifier~\cite{dafny-ai} demonstrated that GPT-4 augmented with
retrieval (RAG) and chain-of-thought prompting synthesises verified Dafny methods for 58\% of
benchmark problems.  More recently, DafnyPro~\cite{dafnypro} achieves 86\% on the DafnyBench
suite using Claude Sonnet~3.5, a 16-point improvement over the prior state of the art.

\paragraph{Our approach: interactive exploration.}
The present work differs from the above in several important ways. Usually, LLMs are used as a black-box component embedded in a fully automated end-to-end pipeline. Our work adopts an interactive, exploratory methodology where a user collaborates with an agent to progressively strengthen specifications. This research’s primary goal is the quality of the resulting specifications, specifically, minimising the trusted base by reducing assumed statements and axiomatic lemmas. Whereas prior systems optimise for automatically closing proofs, we deliberately study the family of structurally related DLL implementations against a shared abstract specification,  exposing what aspects make this data structure hard to verify. 

\subsection{Doubly Linked Lists}
\label{sec:intro:dll}
The investigation in~\cite{dll-paper} presents how to implement a DLL in Rust while satisfying performance requirements and the absence of security vulnerabilities. The focus of the article is on finding the most balanced DLL design when Rust’s ownership rules are taken as a hard constraint. The paper contributes (i)~a trait-based abstract
specification shared by all implementations, (ii)~a taxonomy of three
memory models, (iii)~eleven concrete implementations, and (iv)~a
comparative evaluation across nine performance and three safety test
scenarios. All artefacts are publicly available at
\url{https://github.com/xTachyon/paper\_doubly\_linked\_lists}.

\begin{table}[H]
\centering

\caption{Informal contracts for the eight DLL operations.
  $L$ denotes the current list; $L'$ the list after the operation;
  $\mathit{id}_{\mathit{new}}$ the fresh node}

\label{tab:specs}
\small
\begin{tabular}{lp{11.85cm}}
\toprule
\textbf{Operation} & \textbf{Contract (informal summary)} \\
\midrule
\code{next($id_t$)} &
  Returns $\mathit{None}$ if $L$ is empty or $id_t = L.\mathtt{last}()$;
  else the unique $id$ s.t.\ $L.\mathtt{prec}(id) = id_t$. \\
\code{prec($id_t$)} &
  Returns $\mathit{None}$ if $L$ is empty or $id_t = L.\mathtt{first}()$;
  else the unique $id$ s.t.\ $L.\mathtt{next}(id) = id_t$.
  (Mutually recursive with \code{next}.) \\
\code{push\_back($v$)} &
  $L' = L \uplus \{id_{\mathit{new}}\}$; new node is last;
  if $L$ was empty it is also first.
  Returns $id_{\mathit{new}}$. \\
\code{push\_front($v$)} &
  $L' = L \uplus \{id_{\mathit{new}}\}$; new node is first;
  if $L$ was empty it is also last.
  Returns $id_{\mathit{new}}$. \\
\code{insert\_after($id_t$,$v$)\!\!} &
  New node inserted immediately after $id_t$;
  if $id_t$ was last, new node becomes last. \\
\code{insert\_before($id_t$,$v$)\!\!\!\!} &
  New node inserted immediately before $id_t$;
  if $id_t$ was first, new node becomes first. \\
\code{delete($id$)} &
  $L' = L \setminus \{id\}$; predecessor and successor pointers of
  neighbours updated; \code{first}/\code{last} updated if needed. \\
\code{search($f$)} &
  Returns some $id \in L$ s.t.\ $f(L.\mathtt{value}(id)) = \mathit{true}$,
  or $\mathit{None}$ if no such node exists. \\
\bottomrule
\end{tabular}

\end{table}

\paragraph{Trait-based specification.}
The specification of DLLs in~\cite{dll-paper} is expressed as a \emph{trait}, making the
required behaviour explicit and independent of any concrete memory
representation.  The central abstraction is an associated type
\code{NodeId}, which is the implementation's identifier for a list node
(a pointer, an integer index, or a map key, depending on the
implementation). 
Each of the eight primary methods is accompanied by a semi-formal
Input/Output/Return contract in the companion paper.
Table~\ref{tab:specs} summarizes the contracts used as the target
specification in our experiments.

Here we give, as examples, two trait-based descriptions: \Cref{spec:next} supplies the contract of the \suc\ method and \Cref{spec:prec} that of \pred.
A key property of these specifications is that the contracts are
\emph{mutually recursive}: the return clause of \suc\ is defined
in terms of \pred, and vice versa. This means correctness cannot
be established for any single operation in isolation; the entire list implementation must be verified as a whole.

\noindent
\setlength{\tabcolsep}{3pt}

\begin{tabular}{ll}
\hspace{-2mm}\begin{minipage}{0.475\textwidth}
\begin{lstlisting}[language=trait, caption=Next, mathescape, label=spec:next,  frame=tb]
next(NodeId $\id_{t}$): NodeId;
Input
  self: the current list
  $\id_{t}$: the target node in self
Output 
  unchanged
Return
  if self.is_empty() or $\id_{t}$ = self.last()
  then $\None$
  else $\id$ s.t. self.prec($\id$) = $\id_{t}$
\end{lstlisting}
\end{minipage}
& 
\hspace{1mm}\begin{minipage}{0.49\textwidth}
\begin{lstlisting}[language=trait, caption=Prec, mathescape, label=spec:prec,  frame=tb]
prec(NodeId $\id_{t}$): NodeId;
Input
  self: the current list
  $\id_{t}$: the target node in self
Output 
  unchanged
Return
  if self.is_empty() or $\id_{t}$ = self.first()
  then $\None$
  else $\id$ s.t. self.next($\id$) = $\id_{t}$
\end{lstlisting}
\end{minipage}

\end{tabular}

\paragraph{Memory models and implementations.}
The eleven implementations divide into three groups based on how nodes
are stored and how \code{NodeId} is represented
(Table~\ref{tab:impls}).

\begin{table}[h]
\centering

\caption{The eleven DLL implementations from~\cite{dll-paper}.}
\label{tab:impls}

\small
\begin{tabular}{llll}
\toprule
\textbf{Group} & \textbf{Implementations} & \textbf{\NodeId{} type}
               & \textbf{Key mechanism} \\
\midrule
Heap  & \code{RawP}, \code{NonNull}, \code{Rc}, \code{Std}
      & raw/smart pointer & Heap allocation; \code{Rc}/\code{Weak} for cycles \\
Array & \code{Index}, \code{Slab}, \code{Handle}, \code{Arena}
      & \code{u32} / wrapper & Vec + free-list; generation counts \\
Map   & \code{HashMap}, \code{BTreeMap}, \code{SlotMap}
      & \code{usize} key & Key-value store; monotone key counter \\
\bottomrule
\end{tabular}

\end{table}

In the heap group, each node is an independently heap-allocated struct
and \code{NodeId} is a raw pointer or a pointer wrapper. The
self-referential structure requires either raw pointers (\code{RawP},
\code{NonNull}) or smart pointers such as \code{Rc} and \code{Weak}.
In the array group, all nodes are stored as \code{Option<Element<T>{}>}
slots in a \code{Vec}; a separate free-list tracks available slots
(\code{Index}), or a third-party \textit{arena} crate manages slot reuse
(\code{Slab}, \code{Handle}, \code{Arena}).
In the map group, nodes are stored as values in a hash map, B-tree map,
or slot map, and \code{NodeId} is a monotonically increasing
key, specific to each map type.
The \code{Std} implementation wraps Rust's standard-library
\code{std::collections::LinkedList} and serves as a performance baseline.

\section{A Progressive Verification Case Study}
\label{sec:verificationCaseStudy}

  The research follows three scenarios, each providing challenges and insights. Identifying the challenges and proposing solutions is the basis of understanding the needs of a Verus proof in terms of syntax and proof reasoning. Each experiment is intended to address the shortcomings of the previous one. 

  The main goal is to create a specification as close as possible to a complete machine-checked verification, minimising the \code{assume} statements, which introduce unproven hypotheses inside a proof or executable block, and axiomatic lemmas---\code{external body} specification functions without a body---defined only by the contract (preconditions and postconditions).

  The DLL implementations that are explored in this section can be found at \url{https://github.com/xTachyon/paper_doubly_linked_lists} and the results of the experiments at \url{https://github.com/andradaAntoneac/Verus-dll-Experiments}.

\subsection{First Experiment - Manual Verification}
The first experiment is an onboarding exercise: we hand-write verification code for the index-based DLL to learn Verus syntax and its proof model. Contracts are deliberately weak, so many correctness obligations, such as node  reachability preservation and view-length updates, are discharged with local \code{assume} statements, leaving key proofs informal. The modeled conditions capture only the most obvious shape constraints, such as index bounds and basic list validity. As a result, the experiment prioritizes generating a working Verus model over expressing the full, precise correctness properties expected from a machine‑checked proof (see \Cref{tab:firstVersionSummary}). Because nodes are referenced by array indices rather than Rust references, Verus cannot automatically relate a mutation at index \code{i} to the abstract list view. Any update of the structure has to be asserted manually and captured in the list \code{view} (the abstract, mathematical representation of the list) synchronization as well. 

\begin{table}[ht]
\centering

\begin{tabular}{p{4.2cm} p{10.8cm}}
\hline
\textbf{Summary} &  \\
\hline
Invariant strength & shape-only (index bounds, basic validity) \\
Assume count & 9 (reachability of every node after an altering operation)\\
Verified / total functions & 16  \\
Axiomatic Lemmas & 3 (valid neighbours of a reachable node are also reachable, the view of the list after a deletion is the old view where the deleted node is skipped, the deletion of a node results in a view length smaller than the old view by 1)\\
Operations covered & \code{new}, \code{allocate}, \code{link}, \code{push\_back}, \code{push\_front}, \code{insert\_before}, \code{insert\_after}, \code{delete}\\
Residual gaps & reachability, view-length, and chain consistency all assumed\\
Takeaway & establishes baseline fluency: identifies aliasing, view sync, and framing \\
\hline
\end{tabular}
\caption{First Specification Version Summary for Index DLL}

\label{tab:firstVersionSummary}
\end{table}

\subsection{Second Experiment - Property-Specific Verification}
The second experiment shifts to enforcing a specific, central property, in this case, the \textit{validity of the chain} in the DLL. The validity of the chain is defined as follows: the \code{first} node must be reachable from the \code{last} node by following \code{prec} links, and the \code{last} node must be reachable from the \code{first} node by following \code{next} links, both within the same length‑bounded traversal. The specification is strengthened with a well-formedness predicate (\code{wf()} is the conventional name) and richer invariants, and the proof burden is shifted into explicit axiomatic lemmas (e.g., chain preservation, update-field stability, node reachability after linking). The number of manual assumptions is still high, as it is still not an ideal version of complete proof, but rather an illustration of where and how a property can be asserted throughout the DLL specific operations. The residual gaps are attributable to the DLL's index-based memory model, which stresses both Rust's ownership discipline and Verus's SMT encoding (see \Cref{tab:secondVersionSummary}).

\begin{table}[ht]
\centering
\begin{tabular}{p{4.2cm} p{10.8cm}}
\hline
\textbf{Summary} &  \\
\hline
Invariant strength & chain-focused (\code{wf()} with reachability, index partitioning) \\
Assume count & 6 (the validity of the chain for intermediate states, the validity of each index in the list)\\
Verified / total functions &  11 \\
Axiomatic Lemmas & 2 (the validity of the chain in both directions based on the fact that the list view remains unchanged)\\
Operations covered & \code{new}, \code{allocate}, \code{link}, \code{push\_back} \\
Residual gaps & missing insert/delete ops; chain preservation in link/push\_back still assumed \\
Takeaway & expands from allocation-only to structural updates while keeping proofs lightweight; still relies on assumptions for chain reasoning \\
\hline
\end{tabular}
\caption{Second Specification Version Summary for Index DLL}

\label{tab:secondVersionSummary}
\end{table}

\subsection{Claude-Assisted Scenario}
The experiment with Claude targets complete formal verification of the index-based DLL implementation against the full \code{full\_wf()} invariant, twelve structural conjuncts covering slot/free-list bookkeeping, ghost-set membership, chain integrity, and the abstract list-view sequence, modelling multiple properties across all eight public operations. 
The Verus specification was generated by Claude, using the trait-based description and 
the implementations as input.

The work evolves through four phases tracked by an \code{assume} statement count that dropped from 75 to 11. The key architectural decisions are a two-level invariant split (\code{wf()} for bookkeeping, \code{full\_wf()} for structure) that decouples the helper primitives (e.g., \code{allocate} and \code{link}) from chain reasoning, and the introduction of an explicit ghost field dll: \code{Ghost<Set<u32>{}>} in place of a computed set definition, which turns opaque lambda queries into first-class SMT propositions. Phase 2 discharges 17 acyclicity \code{assume} statements by adding explicit conjuncts and five new lemmas. Phase 3 eliminates a further 44 \code{assume} statements through the ghost boolean pattern (capturing link postconditions in named ghost booleans before they leave Z3's local context) and strengthens \code{allocate} postconditions. Phase 4 closes 3 more via direct witnesses. The final state is 45 verified, 0 errors, 11 assumes, with the residual 11 all sitting at the chain-to-sequence bridge, connecting the exact-hop-count \code{chain\_n} predicate to the abstract \code{list\_view()} sequence index, which is the one structural gap neither the lemma library nor Claude is able to close within the experimental scope. The encountered challenges are intrinsic to the \emph{problem domain} (verification of DLLs in Rust). However, the main problematic aspect is the AI agent's prolonged reasoning time (a couple of hours) and the need for the user to manually 
prompt the agent about aspects it had omitted from the proof.

\subsection{Main Challenges}
\label{sec:mainChallenges}

After experimenting with multiple approaches in the above described scenarios, we can formulate the main challenges of the Verus-based verification for DLLs.  They are split into three categories: challenges related to the Verus syntax and proof structure, challenges intrinsic to the DLLs' particularities and memory models, and the ones resulting from the combination of the abstract model and the technology. The challenges are formalised in collaboration with Claude, as most of them have been encountered during the multiple generation sessions. Some of them were suggested by Claude himself, and the authors edited and checked them.

\subsubsection{Challenges Raised by Using Verus in the Verification of Rust Implemementations}
\label{sec:challenges:verus}

\paragraph{Limited support for raw-pointer implementations.}
The \code{delete} operation and the raw-pointer implementations use Rust's
\code{unsafe} keyword.  Verus can verify \code{unsafe} code in principle, but
the required annotations (pointer validity predicates, aliasing restrictions,
layout proofs) add significant overhead and interact with the existing ghost
infrastructure in non-trivial ways, leaving this kind of support incomplete.

\paragraph{SMT trigger management.}
Verus compiles its annotations into SMT queries discharged by Z3. Z3 reasons efficiently about ground facts but is highly sensitive to how universally quantified propositions are triggered.  A set defined as $\{i \mid \mathit{data}[i].\mathtt{is\_some()}\}$ is opaque to Z3 unless a trigger fires on a membership query.  In practice, every universally quantified invariant conjunct requires careful placement of \code{\#[trigger]} annotations to ensure Z3 instantiates it at the right
proof sites. If a trigger’s specification is missing, then Z3 chooses one that might not be of any help in the proof. Incorrect triggers produce silent failures, as the invariant is logically true, but Z3 cannot see it.

\paragraph{Ghost state ordering and \texttt{old(self)}/\texttt{final(self)} semantics.}
Verus's \code{old(self)} always refers to the function entry state. In a multi-step operation such as \code{push\_back} (\code{allocate} then \code{link} then an assignment),
the postconditions of \code{allocate} are removed from Z3's context when the next mutation is issued.  A \code{let ghost snap = self.data@}
immediately after each mutating call is required to preserve intermediate
states.  Omitting a snapshot produces an error that may manifest several
proof lines later with a message that is not obviously connected to the
missing snapshot.  This issue is pervasive in any multi-mutation operation
and constitutes a significant cognitive overhead for AI-generated proofs,
which must explicitly plan the ghost snapshot sequence.

\paragraph{Mutable references as function parameters.}
A closely related pitfall arises whenever a helper function receives a
mutable reference (\code{\&mut}) to the structure. The prover treats the call
as an opaque black box and assumes that \emph{any} field may have been
arbitrarily modified, retaining only the facts explicitly guaranteed by the
callee's postconditions. Consequently, the preservation of fields that the
function does not intend to change must be stated explicitly in its
\code{ensures} clause (e.g., \code{final(self).other\_field == old(self).other\_field}).

\paragraph{Proof block placement relative to \code{exec} assignments.}
Executable assignments that update \hfill \break \code{self.first}, \code{self.last}, and
\code{self.dll} may pass through intermediate states that violate
\chainwf{}. Establishing the required facts about these transient states
demands proof blocks placed at exactly the right point in the statement
sequence, a placement that is not always easy to guess. A misplaced
\code{proof} block asserts a property of the wrong intermediate state, and
Verus reports only the final postcondition failure, giving no hint that the
block itself is the cause.

\paragraph{Recursive spec function opacity.}
Verus's spec functions are not unfolded automatically by Z3 beyond a finite
recursion depth.  For the predicates that are defined by structural recursion on a fuel parameter, Z3 will not unfold them without an explicit proof lemma that provides the
inductive step. This means every use of such a predicate at a specific
depth requires a lemma invocation. 

\paragraph{Bitvector and mixed-arithmetic casts.}
Verus distinguishes \code{u32}, \code{nat}, and \code{int} as separate types
and does not automatically insert casts.  Subtraction of two \code{nat}
values produces an \code{int} (since the result could in principle be
negative), so an expression such as \code{old\_pos[i] - 1nat} has type
\code{int} and is rejected by a \code{Map<u32,~nat>} value function.
The fix is \code{(old\_pos[i] - 1) as nat}, but this cast is only sound
when \code{old\_pos[i] >= 1}, which itself requires a proof.

\paragraph{Z3 resource limits and context size.}
As the proof file grows,
Z3's default resource limits become binding.  Functions with large proof
blocks (e.g., with many cases and multiple lemma calls
per case) occasionally time out.  The symptom is a resource-limit error rather
than a clear proof failure, making it hard to distinguish ``proof is wrong''
from ``proof is right but Z3 ran out of time.''  Decomposing large proof
blocks into smaller helper lemmas, or adding \code{proof \{ \}} sub-blocks
to give Z3 smaller sub-goals, mitigates this but further increases the number
of proof artefacts to maintain.

\subsubsection{Challenges Raised by the Specification and Verification of DLLs
            Implemented in Rust}
\label{sec:challenges:dll}

\paragraph{Self-referential structure and Rust's ownership model.}
A DLL is \emph{inherently} self-referential, meaning each node holds identifiers to its two neighbours, a mutual borrow that safe Rust forbids. The eleven implementations of~\cite{dll-paper}
each circumvent this constraint by a different mechanism: raw pointers,
\code{NonNull}, \code{Rc<RefCell<T>{}>}, integer indices, or map keys.
Every workaround displaces the self-referential burden from the type checker
to the verifier.  For raw-pointer implementations, the verifier must reason about pointer provenance and disjointness invariants that current verification tools support only partially. For the index-based implementation studied here,
the verifier must track which slots are live, which belong to the free list,
and which are reachable from \code{first}, three overlapping but distinct
concepts that must be kept in sync.

\paragraph{Inherently global invariants.}
The DLL invariant cannot be decomposed into local per-node properties.
Whether a node $v$ is ``valid'' depends on the full list structure: its
\code{prec} must equal the \code{next} field of the preceding node, which
in turn depends on the preceding node's position in the chain, and so on.
This global character means that every invariant conjunct refers to the underlying memory model and every proof step must account for the global effect of even a single link update.

\paragraph{Specification imprecision.}
The semi-formal contracts of~\cite{dll-paper} are adequate for communication
but not for machine-checked proof.  The imprecision gap between documentation-level and verification-level specifications must be filled \emph{before} starting the proof.

\paragraph{Memory model diversity across eleven implementations.}
The three memory model groups (heap, array, map) each require a distinct
ghost infrastructure.  
For example, the predicates expressing the chaining properties are portable, 
but the surrounding invariants are not.
Heap-based implementations require pointer disjointness and ownership
invariants. Map-based implementations require key-monotonicity and absence
of hash collisions.  Sharing the chain lemma library across groups is
possible (as Section~\ref{sec:skill} proposes), but each group's
\emph{wiring} is unique.  A verification effort covering all eleven
implementations cannot avoid rebuilding the ghost infrastructure three times, once per group, and then specialising per implementation.

\paragraph{Absence of a deallocation model.}
The abstract specification has no model for memory recovery. The specification does not constrain what ``deletion'' means in terms
of the abstract view. It specifies the list contents but not the physical
memory layout after the operation.  This simplification keeps the
verification tractable, avoiding separation-logic-style capabilities, but it leaves the
implementation's space behaviour unverified.  Any extension of the
specification to include memory safety or space complexity would require a
substantially more powerful specification framework.

\paragraph{Minimal postconditions for primitives.}
A function like \code{allocate} cannot prove \wf{} for the recycling path without a
no-duplicates property on \code{free\_list}, which is not in \wf{}. Treating this correctly means weakening the primitive’s postconditions. The primitive should only promise facts that every caller can rely on, and avoid baking in stronger claims that require extra context. In other words, the challenge is to avoid over‑specifying a low‑level operation. The solution is to push stronger guarantees upward to the callers that actually have the necessary invariants.
\subsubsection{Combined Challenges Raised by the Verification of DLLs Using Verus}
\label{sec:challenges:dllVerus}

The two preceding subsections catalogued  Verus-tool challenges
(\Cref{sec:challenges:verus}) and DLL-domain challenges
(\Cref{sec:challenges:dll}) independently.  In practice the two sets
do not compose additively, but mostly each domain challenge amplifies one or more
tool challenges, producing a combinatorial burden that exceeds the sum of
its parts.  This subsection describes the concrete interaction pairs
observed in our experiments.

\paragraph{Global DLL invariants~$\times$~SMT trigger management.}
Every conjunct of \fullwf{} quantifies universally over all live nodes,
so each must carry a trigger that Z3 can instantiate at every relevant
proof step. A membership anchor such as \code{\#[trigger] dll@.contains(i)}
suffices for simple conjuncts, but several conjuncts additionally reference
\code{data@[i as int]} or \code{suffix\_pos@[i]}, requiring compound
triggers. Across the conjuncts of \fullwf{} and \chainwf{}, this yields a
dense trigger-annotation discipline. The dangerous failure mode is not a
missing trigger, which Verus usually reports, but a trigger that is
well-formed yet too specific, so Z3 never fires it and the conjunct is
never instantiated. Because the resulting failure surfaces as an unprovable
\code{ensures} rather than any trigger diagnostic, one must distinguish
``conjunct not implied'' from ``conjunct not instantiated'', a distinction
Verus's error messages do not surface, and which is recoverable only by
inspecting quantifier-instantiation logs.

\paragraph{Self-referential mutation~$\times$~ghost-state ordering and
proof-block placement.}
Each mutating DLL operation performs some physical sub-steps
(e.g., \code{allocate} a slot and \code{link} it into the chain), and the global invariants
hold only after \emph{all} sub-steps complete.  Verus's
\code{old(self)} semantics captures state at the function boundary, not
at intermediate sub-steps, so intermediate states require explicit ghost
snapshots and proof blocks for the preconditions of the sub-steps.  

\paragraph{Specification imprecision~$\times$~ghost-state ordering.}
Specification gaps discovered during Verus proof derivation compound the
multi-step ordering challenge.  
Wherever the abstract spec underspecifies a precondition, the Verus
translation must choose a concretisation, and that concretisation
interacts with the ghost-snapshot discipline.

\paragraph{Mutually recursive specification~$\times$~recursive spec
function opacity.}
The abstract DLL specification expresses \code{next} and \code{prec} as
mutually dependent operations (\Cref{sec:intro:dll}).  In Verus,
these translate to the pair
\code{valid\_next\_chain}/\code{valid\_prec\_chain}, both defined as
recursive spec functions.  Because Verus's opacity rules prevent Z3
from automatically unfolding either function, every proof step that
references the forward or backward chain requires an explicit lemma
invocation.  The mutual dependency doubles the required lemma library, as
every structural property (frame, monotonicity, one-step extension,
transitivity) must be proved independently for both directions.

\paragraph{Fuel bound specification~$\times$~reachability property.}
"Reachability” is essential to show that links do not skip or cycle and
\code{valid\_next\_chain} is the right statement for
``the list is bounded-reachable''.  It is not strong enough for inductive
chain lemmas, which need the exact step count for their termination argument. Another function (in our case
\code{chain\_n}) provides that count.  For inductive chain
reasoning in SMT, both a fuel-bounded reachability predicate and an
exact-count predicate should be maintained.

\section{Towards a Verification Skill for DLLs}
\label{sec:skill}

A skill is a package of knowledge, context, and procedural guidance that an AI/LLM agent can load on demand, whenever the situation calls for it. It is usually perceived as a briefing document that the agent reads before tackling a specific kind of task, which replaces the long explanations in the prompts.

At its core, a skill is just a folder. It has a file that acts as an orchestrator. It presents what the skill is for, how to approach the task, what tools to use, what common mistakes to avoid, and how to handle errors. The skill can be accompanied by any other supporting materials, such as reference documents with domain-specific knowledge, scripts that automate parts of the workflow, and templates or other assets the agent might need.

The main purpose of a skill is to turn a general-purpose agent into a specialist. A well-written skill loads exactly the knowledge the agent needs, instead of forcing the agent to gather that information itself, a process that is often error-prone.

The experience accumulated during the experiments and the challenges encountered led to the following set of principles that the skill should follow when generating a Verus based verification of a DLL.

\subsection{Principles Guiding the Skill Design}
\label{sec:skillPrinciples}
\paragraph{Separation of concerns.}
The skill isolates analysis, mapping, generation, verification, and debugging into distinct phases to avoid conflating specification design with proof tactics or diagnosis. 
The analysis refers to input verification, confirming the abstract specification (the starting point of the proof) and the actual implementation are provided and suitable to proceed. Between these two a mapping is drawn, meaning the operations modeled by the specification are identified in the implementation. Mapping serves as the guide for Verus specification generation, which is verified and debugged.
This modularity supports clarity and repeatability. 
The workflow details can be found in \Cref{subsec:aFirstVersionOfTheSkill}.

\paragraph{Traceability.}
The abstract specification is translated into Verus specification code following the rule that every clause (preconditions defined in \code{requires} blocks, postconditions from \code{ensures} blocks, \code{decreases} statements, and \code{invariants}) is justified by a specific abstract requirement. This rule prevents arbitrary or speculative postconditions.

\paragraph{Soundness over convenience.}
Weakening specifications to help the verifier or to generate the specification faster should be forbidden. Bugs might still be found in the implementations, and the proper step after finding one is to report it first, rather than modelling the specification around the bug.

\paragraph{Reusable proof patterns.}
To avoid ad hoc reasoning for commonly used \code{proof} blocks, their structure has to be shaped beforehand.  The agent should have knowledge about the structure of inductive proofs, lemma application, and trigger tuning, for example.

\paragraph{Boundary discipline.} 
As per the aforementioned Verus limitations, the support for \code{unsafe} blocks, raw pointers and opaque library types is present but incomplete and insufficient for this kind of proof. The verification must switch to an axiom‑boundary strategy in this context and document it.

\paragraph{Structure preservation.}
The structures used in the implementation must be preserved as much as possible, so the generated specification proofs are not modeled over surrogate structures. Ghost fields may be used as a means to help the proving process.

\paragraph{Coverage of behaviour.}
The skill requires as input a target property to be proven for the list operations, but that does not mean the operations’ own functional behavior can be ignored. For example, in the \code{push\_back} scenario, there must be a postcondition to verify that the last node has been updated. To avoid under-specification, the property description must contain full functional behaviour modelling.

\subsection{A First Version of the Skill}
\label{subsec:aFirstVersionOfTheSkill}
The skill is used with an LLM agent. It requires a source file for an actual DLL implementation and at least one target property that will be verified for each list operation.
\paragraph{Structure.} The skill is organised into the following components:
\begin{description}
    \item[Core Definitions] The main file and the entry point of specification generation, \code{SKILL.md}, acts as the orchestrator of the generation's behaviour. It formalises the workflow of the skill, including required inputs, mandatory constraints, and non‑negotiable verification rules.
    \item [Property Module] This module contains the descriptions of proof logic for common properties of DLL. At this moment, it defines the correctness property for reachability‑based chain validity, along with proof obligations and failure modes.
    \item [Reference Library] The \textit{Knowledge Base} refers to multiple aspects needed for the demonstrations and generation: the semi-formal abstract description of the DLL, the formal syntax of Verus constructs, \code{spec} and \code{proof} patterns, a list of common pitfalls and remediation steps, a decision framework for interpreting failed proofs (weak spec, mismatch, bug), and a formalisation of when preconditions permit specification simplification, usually when preconditions cancel out some parts of the executable code.   
    \item [Templates] The templates standardize the structure of function contracts, abstract structure specifications, proof skeleton for target properties, how to construct bug reports that are complete, actionable, and comparable across cases..
    \item [Examples] This module is used for showcasing how to tackle situations that might arise, such as demonstrating permissible structural simplification without loss of semantic fidelity or correct classification and reporting of implementation defects.
    \item [Procedural Orchestration] The heart of the skill is the pipeline that consumes the above assets. The skill uses it as the \textit{workflow}, and it is detailed below.
    
\end{description}
\paragraph{Workflow.}

The following main steps are intended to implement the aforementioned design principles for addressing the verification challenges:
\begin{enumerate}
    \item Analyse the inputs: In this step, the agent builds a mental model of the implementation and abstract specification. The main goal of this step is to identify functions, data types, integer types, recursions, loops, and the implementations of the functions found in the abstract specification: \code{first}, \code{last}, \code{next}, \code{prec}, \code{value}, and to decide whether the code is an axiom boundary. The output is a structured summary of the implementation referring to functions, types, preconditions, postconditions, invariants, and target properties.
    \item Map abstract specification to Verus: This step translates the abstract specification into Verus constructs. Any semi-formal specification is mapped to a proper Verus statement, such as \code{requires}, \code{ensures}, \code{spec fn} (marking a function as specification only), \code{forall}, etc. This step is also responsible for defining the storage‑agnostic DLL model: how node identification should be defined in Verus, how node membership is modeled, how first/last nodes are structured. 
    \item Generate the specification: Starting from the mental model and the mapping of the abstract specification to Verus syntax and structure, a first specification is generated. This step has the most workload. If the first step classified the implementation as axiom-boundary, a separate strategy is defined: executable functions are marked with \code{external\_body}, code bodies should not be proved, and the focus is on defining precise preconditions and postconditions in the requires and ensures blocks and on consistency lemmas. Otherwise, the flow is to generate the \code{requires} and \code{ensures} blocks and define the \code{spec fn} helpers. Several templates are provided in this step for function definition and structure specification.  An optional inter-step is to simplify unreachable branches under strong preconditions. The agent is guided to encode precise link updates with frame clauses. For properties for which we expect a specific proof, such as the validity of the chain, the skill guides the agent to generate a specification using the already defined property description.
    \item Add invariants: For every loop or recursion, invariants and a \code{decreases} clause must be provided. For traversal loops, reachability, visited‑set, and progress conditions should be included. This is mandatory whenever loops or recursion appear and functions primarily as a verification checkpoint.
    \item Verify property: Another verification step is to check that the target property is ensured through the postconditions and the actual function code. The \code{proof} or \code{assert} blocks express the property at the end of a function if the operation is complex. The skill guides the agent to use intermediate proofs or temporary \code{assume} statements. The main goal is to ensure each executable function's \code{ensures} are discharged by the actual code (no modelling-only specification).
    \item Debug failures: This step classifies the failure by error message and applies targeted fixes, which might be strengthening invariants, adding bounds to prevent overflow, splitting postconditions, or adding intermediate lemmas. It also addresses common pitfalls in Verus specifications and syntax, where straightforward fixes are provided.
    \item Report discrepancies: If the failure is a real bug or spec mismatch, the agent must produce a structured discrepancy report with a concrete counterexample, the violated abstract requirement, and a suggested fix. The skill strengthens the rule to never weaken the specification to hide an implementation defect.
\end{enumerate}

\paragraph{How the Challenges are Addressed by the Skill.}
\label{sec:addressedChallenges}
Some challenges presented in \Cref{sec:mainChallenges} are addressed directly through a concrete skill specification, whereas others are addressed by the skill as a whole. 

The self-referential nature of DLLs in formal verification implies rebuilding the ghost infrastructure for each implementation. Rather than proving the list’s internal structure, the skill guides the agent toward a storage-agnostic representation of the list. The node identification and membership are abstracted away from the storage mechanism, so the proof focuses on the global DLL properties. This specification also addresses the memory model diversity across eleven implementations. 

The traceability principle ties each invariant clause to an abstract requirement to address the inherently global invariants challenge: the global effects of a single link update determine the node’s validity. The principle also prevents arbitrary postconditions, reducing imprecision.

The separation of concerns isolates proof tactics and debugging into distinct phases.  The verification and debugging phases address possible proof-block misplacement. The agent is instructed to identify operations and sub-steps that interfere with proof-block placement in a complex demonstration.

\subsection{Current Results and Resolutions}
\label{sec:currentResultsAndResolutions}
The current version of the skill is not a final one. To observe the shortcomings of the skill and how they might be addressed, multiple experiments should be conducted and analysed over multiple iterations.

The evaluation of a skill's results depends on its own design, meaning that the user is responsible for analyzing the different agents' generated Verus code against the proposed workflow and checking whether the specifications of the skill are satisfied and how. 

\Cref{tab:comparisonOfGeneratedDemonstration} presents some examples of generated specifications with different agents. The index-based implementation is the main use-case as it provided the most insights for the skill construction (I1, I2, I3, I4 rows represent different runs for the index implementation), whereas the Standard DLL (STD row) and SlotMap DLL (SM row) represent more of a testing step. The I2 and I3 experiments refer to the same implementation and use the same agent, but are generated using different prompts. All the experiments are conducted using the prompt \code{/generate-verus-specification  for <path to the implementation> targeting the valid chain property}. However for the I2 experiment it is also specified that \textit{"a full demonstration is wanted"}.

\vspace{2pc}
\begin{table}[ht]
\centering
\begin{tabular}{l l c c c c c}
\hline
Acronym & Implementation & Model & Assumes & Axioms & Code lines & Time\\
\hline
I1 & Index & Codex 5.2 High & 46 & 0 & 612 & 10 min.\\
I2 & Index & Claude 4.8 Max & 0 & 0 & 2259 & 50 min.\\
I3 & Index & Claude 4.8 Max & 0 & 0 & 1539 & 40 min. \\
I4 & Index & Claude 4.6 High & 18 & 1 & 1122 & 20 min.\\
STD & Standard & Claude 4.6 High & 0 & 8 & 485 & 10 min.\\
SM & SlotMap & Claude 4.6 High & 0 & 8 & 762 & 10 min.\\
\hline
\end{tabular}
\caption{Generated Demonstration}
\label{tab:comparisonOfGeneratedDemonstration}
\end{table}

\subsection{Specifications' Evaluation Criteria}

The skill is designed so that it guides the verification-code generation rather than constrains the proof methods. It gives steps to follow and does not enforce all the details of the verification, as there would be no reason to use an agent for this task. 

The following criteria are used to capture the choices each agent made following the same workflow, and the resolutions for each can be observed in \Cref{tab:perSpecificationEvaluationCriteria}.

\begin{enumerate}
    \item Soundness boundary: where verification stops and trust begins (explicit \code{assume} and \hfill \break \code{external\_body}).
    \begin{itemize}
        \item None: no \code{assume} or \code{external\_body}; code is fully proved.
        \item Narrow: a small isolated trusted helper; core mutators proved.
        \item Moderate: mixed boundary; some mutators rely on assumptions.
        \item Full/opaque: most or all mutators are \code{external\_body}.
    \end{itemize}
    
    \item Spec alignment: how much of the chain validity demonstration is followed.

    \begin{itemize}
        \item Direct: \code{valid\_chain} states reachability from first and to last.
        \item Witness-based: \code{valid\_chain} uses a ghost order/chain, and reachability is derived as a lemma.
        \item Traversal-derived: \code{valid\_chain} defined via \code{chain\_seq}/\code{traverse} coverage, reachability is a corollary.
        \item Axiomatic: reachability is assumed via postconditions, not computed.
    \end{itemize}

    \item Order precision: how exactly postconditions pin down relative order changes.
    \begin{itemize}
        \item High: exact sequence equations (append/prepend/remove).
        \item Medium: order witness exists but not all ops specify positions.
        \item Low: only membership/endpoint facts (e.g., node set + head and tail nodes).
    \end{itemize}
    
    \item Generality: how restrictive the spec is in type bounds or modelling choices.
    \begin{itemize}
        \item Unconstrained: no \code{T} bounds; minimal extra state.
        \item Type-bounded: requires \code{T: Copy}.
        \item Ghost-heavy: adds ghost fields like order or links.
        \item Representation-bounded: relies on sentinel/capacity assumptions.
    \end{itemize}
    
    \item Allocator modelling: how allocation and reuse correctness is specified.
    \begin{itemize}
        \item Invariant-level: \code{free\_list\_wf} (or equivalent) in the global invariant.
        \item Per-operation: each mutator assumes allocator well-formedness.
        \item Implicit: allocator behaviour not modeled.
        \item Opaque/freshness axioms: uniqueness/freshness assumed for an opaque store.
    \end{itemize}
        
    \item Proof ergonomics: where proof effort lives (lemma toolkit vs ghost instrumentation).
    \begin{itemize}
        \item Lemma-heavy: many inductive lemmas, minimal ghost state.
        \item Ghost-heavy: extra ghost state simplifies proofs and reduces lemmas.
        \item Balanced: moderate lemmas and moderate ghost state.
        \item Axiom-driven: few lemmas; behaviour mostly encoded in postconditions.
    \end{itemize}
\end{enumerate}

\begin{table}[ht]
\centering
\begin{minipage}[t]{0.48\linewidth}
\centering
\renewcommand{\arraystretch}{1.1}
\begin{tabular}{|p{0.55\linewidth}|p{0.35\linewidth}|}
\hline
\multicolumn{2}{|c|}{\textbf{1. Index with Codex 5.2 High}} \\ \hline
Soundness boundary & Moderate \\ \hline
Spec alignment & Direct \\ \hline
Order precision & Low \\ \hline
Generality & Type-bounded (\code{T: Copy}) \\ \hline
Allocator modelling & Implicit \\ \hline
Proof ergonomics & Axiom-driven \\ \hline
\end{tabular}
\end{minipage}\hfill
\begin{minipage}[t]{0.48\linewidth}
\centering
\renewcommand{\arraystretch}{1.1}
\begin{tabular}{|p{0.55\linewidth}|p{0.35\linewidth}|}
\hline
\multicolumn{2}{|c|}{\textbf{2. Index with Claude 4.8 Max}} \\ \hline
Soundness boundary & None \\ \hline
Spec alignment & Direct \\ \hline
Order precision & Low \\ \hline
Generality & Unconstrained \\ \hline
Allocator modelling & Invariant-level \\ \hline
Proof ergonomics & Lemma-heavy \\ \hline
\end{tabular}
\end{minipage}
\vspace{10pt}
\noindent
\begin{minipage}[t]{0.48\linewidth}
\centering
\renewcommand{\arraystretch}{1.1}
\begin{tabular}{|p{0.55\linewidth}|p{0.35\linewidth}|}
\hline
\multicolumn{2}{|c|}{\textbf{3. Index with Claude 4.8 Max}} \\ \hline
Soundness boundary & None \\ \hline
Spec alignment & Witness-based \\ \hline
Order precision & High \\ \hline
Generality & Ghost-heavy \\ \hline
Allocator modelling & Invariant-level \\ \hline
Proof ergonomics & Ghost-heavy \\ \hline
\end{tabular}
\end{minipage}\hfill
\begin{minipage}[t]{0.48\linewidth}
\centering
\renewcommand{\arraystretch}{1.1}
\begin{tabular}{|p{0.55\linewidth}|p{0.35\linewidth}|}
\hline
\multicolumn{2}{|c|}{\textbf{4. Index with Claude 4.6 High}} \\ \hline
Soundness boundary & Moderate \\ \hline
Spec alignment & Traversal-derived \\ \hline
Order precision & High \\ \hline
Generality & Unconstrained \\ \hline
Allocator modelling & Per-operation \\ \hline
Proof ergonomics & Lemma-heavy \\ \hline
\end{tabular}
\end{minipage}
\vspace{10pt}
\noindent
\begin{minipage}[t]{0.48\linewidth}
\centering
\renewcommand{\arraystretch}{1.1}
\begin{tabular}{|p{0.55\linewidth}|p{0.35\linewidth}|}
\hline
\multicolumn{2}{|c|}{\textbf{5. Standard DLL with Claude 4.6 High}} \\ \hline
Soundness boundary & Full/opaque \\ \hline
Spec alignment & Witness-based \\ \hline
Order precision & High \\ \hline
Generality & Ghost-heavy \\ \hline
Allocator modelling & Opaque/freshness \\ \hline
Proof ergonomics & Axiom-driven \\ \hline
\end{tabular}
\end{minipage}\hfill
\begin{minipage}[t]{0.48\linewidth}
\centering
\renewcommand{\arraystretch}{1.1}
\begin{tabular}{|p{0.55\linewidth}|p{0.35\linewidth}|}
\hline
\multicolumn{2}{|c|}{\textbf{6. SlotMap with Claude 4.6 High}} \\ \hline
Soundness boundary & Full/opaque \\ \hline
Spec alignment & Witness-based \\ \hline
Order precision & High \\ \hline
Generality & Ghost-heavy \\ \hline
Allocator modelling & Opaque/freshness \\ \hline
Proof ergonomics & Ghost-heavy \\ \hline
\end{tabular}
\end{minipage}
\caption{Per-specification Evaluation Criteria}
\label{tab:perSpecificationEvaluationCriteria}
\end{table}

\subsection{Shortcomings Yet to Be Addressed}
Each experiment result in \Cref{tab:perSpecificationEvaluationCriteria} has its strengths and shortcomings. It is important to observe them as they guide the development of the skill and the fine-tuning of the specification.

The first two specifications adopt a reachability-based definition of chain validity. I1 relies on numerous explicit \code{assume} statements, which weakens the verification soundness. I2 uses a comprehensive lemma toolkit, including fuel monotonicity, frame conditions, and reachability chaining, and it addresses allocator correctness through a \code{free\_list\_wf} specification included in the state invariant.  The specification lacks a positional witness for ordering, which limits the ability to state or prove fine‑grained order‑preservation postconditions.

I3 presents a different approach from the other three that model the same implementation. This approach strengthens the invariant by introducing a ghost order sequence and proving that the invariant implies reachability. It thus achieves full verification while enabling high‑precision ordering postconditions. However, a ghost-heavy specification is less faithful to the concrete implementation.

A less satisfying result is I4: \code{link}---one of the most used functions, responsible for updating the two adjacent node identifiers between nodes, so in short, creating the list structure---is treated as an \code{external\_body}. The proof is incomplete in the most mutation‑critical parts.

For the Standard implementation, STD uses a ghost chain of heap addresses and axiomatizes all pointer‑touching operations, proving only consistency lemmas and simple accessors. The specification is precise about chain shape but intentionally treats the concrete data structure as opaque. As the actual list implementation is not our creation, the verification is entirely axiomatic with respect to the concrete code and it cannot expose implementation bugs or justify allocator freshness beyond trust assumptions. 

The SlotMap verification is a similar case. The implementation code is not responsible for the memory management of the list, but rather of the list abstraction over a map structure. The specification offers a stronger abstract model than the pointer case despite the opaque SlotMap backend,  enabling bidirectional link‑consistency lemmas while retaining a chain witness.

After seeing these results, the skill might need:
\begin{itemize}
    \item Reduced assumptions footprint by systematic lemma coverage: Even if the skill instructs the agent that the number of \code{assume} statements should be minimal, an intermediate step that instructs the agent to generate the missing lemmas that replace assume stubs should be added.
    \item A post‑mutation proof staging pattern: Many \code{assume} uses arise because facts established before a mutation are not re‑established after the mutation. The skill needs a disciplined staging pattern, adding ghost snapshots after each state change and local lemmas to re‑assert invariants.
    \item Ordering precision:  The skill should either add a ghost order witness or enrich postconditions with sequence equations, as ordering is, alongside reachability, one of the most useful properties to establish for chain validity. 
\end{itemize}

\section{Conclusions}
\label{sec:conclusions}
In this paper, we investigated whether an LLM agent can efficiently verify a family of structurally related Rust implementations against a shared abstract specification, and whether that process can be codified into a reusable methodology. Our progressive case study shows that a key difficulty of verifying Rust programs in Verus lies in specification quality: weakened specifications are masked by assume statements and by axiomatic lemmas built on unverified (trusted) functions that nonetheless report as verified. The proposed skill addresses some challenges leading to this situation.   
The main contribution is the skill and its interaction with different agents. By packaging domain knowledge and a soundness-first workflow into a reusable skill, a general-purpose agent is turned into a DLL-verification specialist, offering a practical path toward LLM-assisted verification whose trusted base is explicit and minimal.
The skill is a first version, and its limitations point to future work: systematic lemma coverage to replace assume stubs, a post-mutation proof staging pattern, and stronger ordering witnesses. 
Our longer-term goal remains verifying all eleven implementations against the single shared specification. This analysis needs to work with different agents and not constrain a single way of proving some established properties. The analysis becomes extensive, but would represent an actual evaluation of the skill and capabilities of a tool such as Verus.

\paragraph{Acknowledgments} The authors warmly thank the anonymous reviewers for many useful suggestions for improvements.

\bibliographystyle{eptcs}
\bibliography{refs}

\end{document}